\documentclass[runningheads]{llncs}
\usepackage[T1]{fontenc}
\usepackage{graphicx}
\usepackage{color}
\usepackage{hyperref}

\usepackage{algorithm}
\usepackage{algorithmic}
\usepackage{amsmath}
\usepackage{multirow}
\usepackage{orcidlink}
\begin{document}
\title{Empirical Characterization of Rationale Stability Under Controlled Perturbations for Explainable Pattern Recognition}
\titlerunning{Rationale Stability Under Controlled Perturbations}
%
%
\author{Abu Noman Md Sakib\inst{1}\orcidlink{0000-0002-0761-035X} \and
Zhensen Wang\inst{1}\orcidlink{0009-0001-3536-4870} \and
Merjulah Roby\inst{2}\orcidlink{0000-0003-3578-8216} \and
Zijie Zhang\inst{1}\orcidlink{0000-0003-1254-098X}}
\authorrunning{A.N.M. Sakib et al.}
%
\institute{
Department of Computer Science, The University of Texas at San Antonio, San Antonio, TX, 78249, USA 
\email{\{abunomanmd.sakib,zhensen.wang,zijie.zhang\}@utsa.edu} \and
Department of Mechanical, Aerospace, and Industrial Engineering, The University of Texas at San Antonio, San Antonio, TX, 78249, USA\\
\email{merjulah.roby@utsa.edu}}
\maketitle              
\begin{abstract}
Reliable pattern recognition systems should exhibit consistent behavior across similar inputs, and their explanations should remain stable. However, most Explainable AI evaluations remain instance centric and do not explicitly quantify whether attribution patterns are consistent across samples that share the same class or represent small variations of the same input. In this work, we propose a novel metric aimed at assessing the consistency of model explanations, ensuring that models consistently reflect the intended objectives and consistency under label-preserving perturbations. We implement this metric using a pre-trained BERT model on the SST-2 sentiment analysis dataset, with additional robustness tests on RoBERTa, DistilBERT, and IMDB, applying SHAP to compute feature importance for various test samples. The proposed metric quantifies the cosine similarity of SHAP values for inputs with the same label, aiming to detect inconsistent behaviors, such as biased reliance on certain features or failure to maintain consistent reasoning for similar predictions. Through a series of experiments, we evaluate the ability of this metric to identify misaligned predictions and inconsistencies in model explanations. These experiments are compared against standard fidelity metrics to assess whether the new metric can effectively identify when a model’s behavior deviates from its intended objectives. The proposed framework provides a deeper understanding of model behavior by enabling more robust verification of rationale stability, which is critical for building trustworthy AI systems. By quantifying whether models rely on consistent attribution patterns for similar inputs, the proposed approach supports more robust evaluation of model behavior in practical pattern recognition pipelines. Our code is publicly available at \url{https://github.com/anmspro/ESS-XAI-Stability}.

\end{abstract}

\section{Introduction}

Ensuring that machine learning models are consistent under label-preserving perturbations is critical for their trustworthy deployment, especially in high-stakes domains such as healthcare, finance, and autonomous systems \cite{han2024trustworthy,li2025scalable}. As machine learning models are increasingly integrated into decision-making processes, the need for transparency and interpretability in these models becomes paramount \cite{qian2024towards,wang2021interpretable}. Explainable AI (XAI) techniques have emerged as powerful tools to address this challenge, providing insights into how models arrive at their decisions \cite{alufaisan2021does}. These techniques aim to make models more transparent, thereby improving user trust and ensuring that AI systems operate in ways that are understandable \cite{agarwal2022openxai}. XAI aims to clarify how machine learning models, particularly deep neural networks, make predictions in order to improve transparency, trust, and reliability in decision-making \cite{pillai2021explainable,carrow2025neural}. Given the complexity and non-linear structure of these models, many existing studies focus on post-hoc explanation techniques that interpret model behavior after training \cite{slack2020fooling,han2022explanation}. These methods include feature attribution \cite{nguyen2021effectiveness} approaches that identify important input features, attention and visualization techniques \cite{hu2023seat} that highlight influential regions in the data, example-based explanations \cite{teso2021interactive} that compare predictions to similar instances, and counterfactual methods \cite{kusner2017counterfactual} that explore how minimal changes to inputs could alter outcomes. Together, these approaches help uncover the internal reasoning of black-box models without modifying their architecture or training process.

In many real-world applications, it is not enough for a model to simply provide explanations for specific predictions \cite{pillai2022consistent}. For instance, if a model provides an explanation for a sentiment classification, the explanation should remain consistent when applied to similar sentences, ensuring that the model’s reasoning aligns with human expectations across a wide range of scenarios. The absence of such consistency verification raises concerns about the reliability of AI systems. While traditional interpretability methods focus on providing explanations for individual predictions, they often overlook the critical aspect of model behavior consistency \cite{moraffah2020causal}. Existing research on XAI primarily emphasizes the generation of local explanations for specific predictions \cite{ghalebikesabi2021locality}, but there is a significant gap in evaluating whether these explanations remain stable when the model encounters similar inputs or variations of the same input. Without addressing this, we risk relying on explanations that may be inconsistent, leading to a lack of trust in the model’s decision-making process. The absence of a systematic way to verify explanation consistency means that models might appear interpretable in isolated instances, but fail to exhibit reliable reasoning across similar cases. This undermines the credibility of AI systems, particularly in applications where consistent decision-making is crucial. Therefore, the need for a robust method to assess and ensure the consistency of explanations is essential for advancing the field of XAI and ensuring that models are not only interpretable but also consistently aligned with their decision-making processes.

To address the issue of inconsistent model explanations, this paper proposes a novel metric that quantifies the consistency of model explanations by leveraging feature attributes. We hypothesize that explanations for similar inputs should exhibit high similarity, and inconsistencies in explanations often arise when the model relies disproportionately on certain features or provides divergent reasoning for similar cases. To capture this, our approach builds upon SHapley Additive exPlanations (SHAP) \cite{shap}, a widely used explanation method, by quantifying the cosine similarity \cite{steck2024cosine} between SHAP values for inputs with the same label. We apply this consistency verification metric to several common environments and evaluate it on benchmark datasets. Our experiments demonstrate that the proposed metric is effective in detecting misaligned predictions and inconsistencies in model explanations. Additionally, we compare the results of our consistency metric with standard fidelity metrics \cite{zheng2025ffidelity,robustfidelity}, providing a comprehensive evaluation of its ability to reveal deviations in model behavior. 
In this work, we propose a practical method to assess and ensure the consistency of model explanations, offering a deeper understanding of model behavior beyond instance-level predictions. The contributions of this work are threefold:
\begin{itemize}
    \item We introduce ESS, a novel metric to quantify explanation consistency across same-label inputs, enhancing model verification for stability.
    \item We present a comprehensive evaluation of ESS on BERT fine-tuned for SST-2, comparing it with fidelity metrics and testing robustness via paraphrasing.
    \item We demonstrate the applicability of ESS in detecting misaligned predictions.
\end{itemize}

Our work offers valuable insights for improving the reliability, trustworthiness, and safety of AI systems in socially sensitive contexts by ensuring that they not only explain their decisions but also do so consistently across similar scenarios. The paper is organized as follows: Section 2 reviews related work. Section 3 describes the overall methodology. Section 4 presents experimental results. Section 5 concludes with key findings and contributions.

\section{Related Work}

Intrinsic interpretability methods aim to design models that are inherently interpretable, such as decision trees, linear models, and attention-based mechanisms \cite{t02,t01}. These models provide a transparent decision-making process but may be limited in their performance when applied to complex tasks such as natural language processing (NLP) and image recognition \cite{chen2019looks}. On the other hand, post-hoc interpretability methods attempt to explain the behavior of black-box models after they have been trained. Some of the most popular post-hoc explanation techniques \cite{slack2021reliable} include Local Interpretable Model-agnostic Explanations (LIME) \cite{lime}, SHAP \cite{shap}, and Saliency Maps \cite{gradcam}. These methods provide feature-level importance scores and visualizations that highlight the key factors driving model predictions. LIME, for instance, generates surrogate interpretable models for individual predictions by approximating the behavior of a black-box model with a locally weighted interpretable model. SHAP, based on cooperative game theory, attributes a portion of the model’s prediction to each feature, providing a unified framework for both global and local interpretability. Grad-CAM, primarily used for convolutional neural networks (CNNs) in computer vision, produces heatmaps that show which regions of an image influence a model’s decision. These methods have been widely adopted in various applications, from text classification to image segmentation, and have proven effective in providing insights into model behavior.

While existing XAI methods offer valuable insights into individual predictions, they often overlook the consistency of these explanations across similar inputs. Ensuring that a model’s explanations remain stable and consistent is crucial for building trustworthy systems, especially in applications where decisions must align with human expectations \cite{han2021explanation}. Several works have started to explore the concept of explanation consistency, but this remains an underexplored area. For example, \cite{fel2022good} proposes methods to algorithmic stability measures of explanations by comparing explanations for different models. However, these methods focus primarily on the stability of feature importance across different models rather than the consistency of explanations for the same model across similar inputs.

The stability of AI models has been a topic of increasing interest, especially with the rise of high-stakes applications. While there has been some work on aligning model outputs with human preferences \cite{kraus2025maximizing}, much of this work focuses on output alignment rather than explanation alignment. The need for consistency in explanations aligns with the broader notion of model accountability and reliability. Recent research \cite{mahmud2024verification} emphasizes the importance of ensuring that machine learning models not only perform well but also provide explanations that are consistent with human values. However, there is still a lack of robust metrics for evaluating explanation consistency across various model architectures and datasets. Our work seeks to fill this gap by introducing a novel metric for verifying the consistency of model explanations. Unlike previous studies that focus on isolated explanations or model fidelity, we propose a method for assessing the consistency of explanations over multiple inputs, offering a more comprehensive approach to model alignment.

\section{Methodology}

\begin{figure*}[h]
\centering
\includegraphics[width=0.8\textwidth]{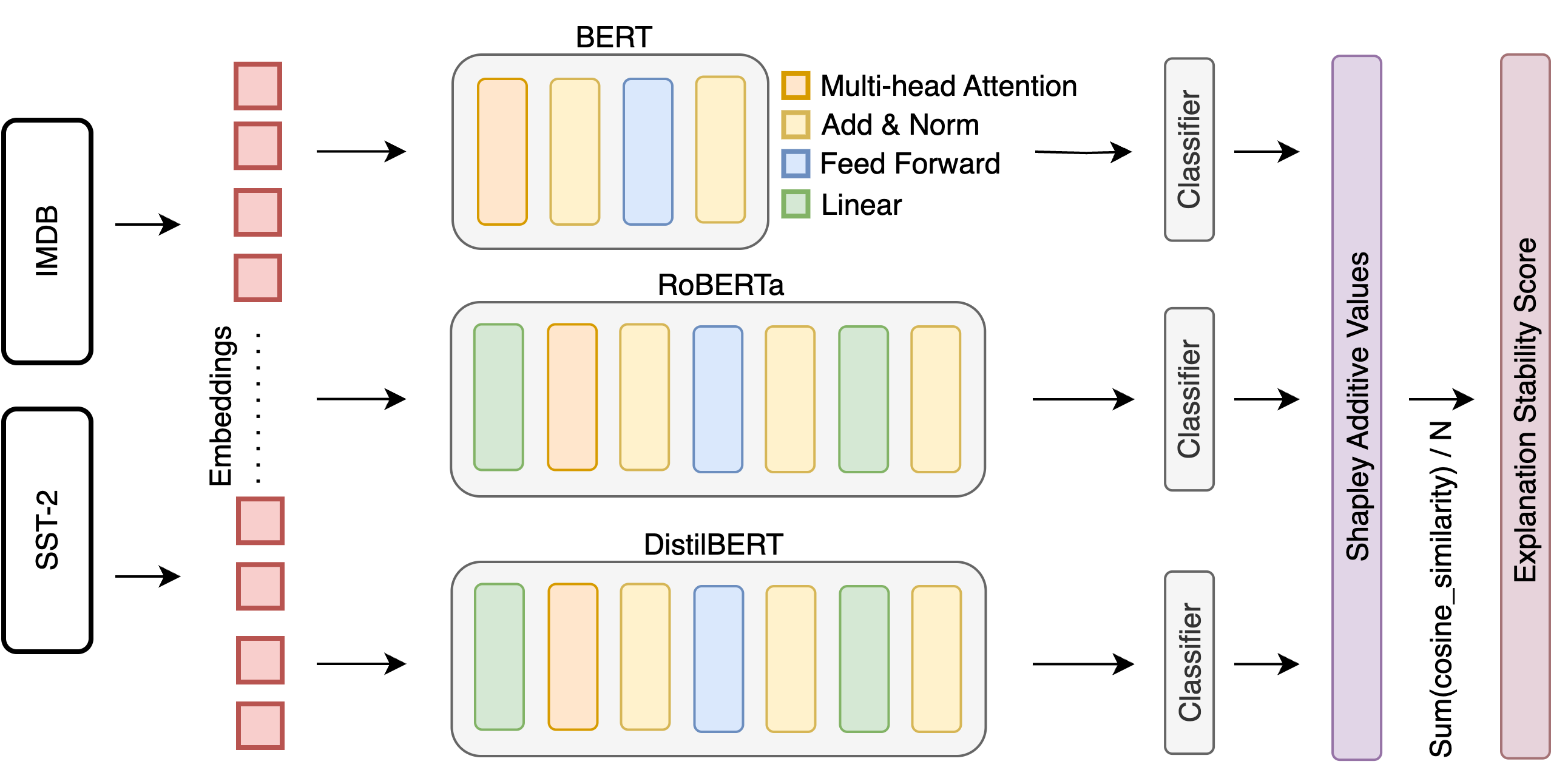}
\caption{Overall methodology for rationale-stability evaluation. Samples from the data are encoded and processed by transformer classifiers, post-hoc token attributions are computed, and the proposed ESS Score is obtained.}
\label{fig:methodology}
\end{figure*}

\subsection{Overview}

In this work, we introduce a novel metric designed to assess the consistency of model explanations. The proposed approach aims to evaluate whether explanations for similar inputs remain stable when a model encounters variations of the same input. The process unfolds in five main steps: first, we set up the pre-trained models and datasets, followed by the generation of model explanations. Next, we define the consistency metric, based on cosine similarity between feature values of inputs with the same label. To assess the robustness of the explanations, we perform an ablation study using paraphrased inputs. Lastly, we introduce evaluation metrics, including the Explanation Stability Score (ESS) and Fidelity measures, to validate the effectiveness of our consistency metric. Figure \ref{fig:methodology} illustrates the whole process.

\subsection{Model Setup and Preprocessing}

For our experiments, we employ several pre-trained models and well-established datasets. The models we use are BERT (bert-base-uncased) \cite{devlin2019bert}, RoBERTa (roberta-base) \cite{liu2019roberta}, and DistilBERT (distilbert-base-uncased) \cite{sanh2019distilbert}, all of which are foundational transformer baseline architectures. These models are fine-tuned on sentiment classification tasks. We utilize two widely recognized sentiment analysis datasets: Stanford Sentiment Treebank v2 (SST2) \cite{sst2}, which is designed for binary sentiment classification, and IMDb \cite{imdb}, a larger dataset that contains movie reviews labeled as positive or negative. We consider the train and test splits when available; for splits without ground-truth labels, we fall back to model-predicted labels for forming same-label pairs. For computational efficiency, we subsample a fixed number of inputs per split. All models are run in evaluation mode, and inputs are tokenized with padding and truncation to a maximum length of 512 tokens using each model’s tokenizer. All random seeds are fixed (42) to ensure reproducibility of subsampling and WordNet-based perturbations. These preprocessing steps ensure that the input data is formatted appropriately for model ingestion and subsequent explanation generation. Algorithm 1 calculates the average cosine similarity between feature values for pairs of inputs with the same label, providing a quantifiable measure of explanation consistency for the model.

\begin{algorithm}
\caption{Algorithm for Consistency Verification}
\begin{algorithmic}[1]
\STATE \textbf{Input:} Models $M = \{\text{bert-base-uncased, roberta-base, distilbert-base-uncased}\}$, Datasets $D = \{\text{sst2, imdb}\}$, Splits $S = \{\text{train, test}\}$, Ablations $A = \{\text{original, paraphrase}\}$
\STATE \textbf{Output:} DataFrame with mean ESS and fidelity (k=5)

\STATE Set random seeds, download WordNet
\FOR{each model $m \in M$}
    \STATE Load tokenizer and model $m$, initialize SHAP explainer
    \FOR{each dataset $d \in D$}
        \FOR{each split $s \in S$}
            \IF{$d = \text{imdb}$ and $s = \text{validation}$}
                \STATE \textbf{continue}
            \ENDIF
            \STATE Load dataset, subsample texts and labels
            \FOR{each ablation $a \in A$}
                \STATE Apply $a$ (paraphrase if applicable)
                \STATE Compute SHAP values and predictions
                \STATE Calculate ESS (cosine similarity) and fidelity (k=5)
                \STATE Append $\{m, d, s, a, \text{mean(ESS)}, \text{mean(fidelity)}\}$ to results
            \ENDFOR
        \ENDFOR
    \ENDFOR
\ENDFOR
\end{algorithmic}
\end{algorithm}

\subsection{Explanation Generation}

To explain the model's predictions, we leverage the SHAP method which interprets model predictions by attributing the output to individual features. For an input \( x \), a model \( f \) produces a prediction \( f(x) \). SHAP decomposes this into contributions from each feature \( f_i \), where the value \( \phi_i(x) \) quantifies feature \( i \)'s impact on the prediction relative to the expected output \( E[f(x)] \).

The SHAP value for feature \( i \) is defined as:
\begin{equation}
    \phi_i(x) = \sum_{S \subseteq \mathcal{N} \setminus \{i\}} \frac{|S|!(|\mathcal{N}| - |S| - 1)!}{|\mathcal{N}|!} \left[ f(x_{S \cup \{i\}}) - f(x_S) \right]
\end{equation}
Here, \( \mathcal{N} \) is the set of all features, \( S \subseteq \mathcal{N} \setminus \{i\} \) is a subset excluding feature \( i \), \( x_S \) is the input with only features in \( S \) active (others set to baseline values), and \( f(x_{S \cup \{i\}}) - f(x_S) \) is the marginal contribution of feature \( i \). The term \( \frac{|S|!(|\mathcal{N}| - |S| - 1)!}{|\mathcal{N}|!} \) is the Shapley weight, averaging contributions across all permutations.

\begin{equation}
    f(x) = E[f(x)] + \sum_{i=1}^{|\mathcal{N}|} \phi_i(x)
\end{equation}
where \( E[f(x)] \) is the expected model output over a background dataset. This ensures the prediction is fully decomposed into feature contributions. For text data, features are tokens or embeddings. SHAP attributes predictions to words or phrases, with \( x_S \) evaluated by masking excluded tokens. Local explanations show how each feature affects a specific prediction, while global explanations, derived from aggregating SHAP values (e.g., mean \( |\phi_i| \)), reveal overall feature importance. Exact Shapley value computation is expensive, requiring \( 2^{|\mathcal{N}|} \) evaluations. Kernel SHAP approximates this via weighted linear regression:
\begin{equation}
    \min_{\phi} \sum_{S \subseteq \mathcal{N}} w(S) \left[ f(x_S) - \left( E[f(x)] + \sum_{i \in S} \phi_i \right) \right]^2
\end{equation}
where \( w(S) \) is the Shapley weight. By calculating the SHAP values, we can generate both global and local explanations, where global explanations offer insights into the model's overall behavior, and local explanations explain specific predictions.

\subsection{Consistency Metric Definition}

The consistency metric quantifies the stability of model explanations. We hypothesize that similar inputs should lead to similar explanations. To capture this consistency, we compute the cosine similarity between the feature value vectors of two inputs \( x_1 \) and \( x_2 \) that belong to the same class. The cosine similarity is defined as:

\begin{equation}
    \text{Cosine Similarity}(v_1, v_2) = \frac{v_1 \cdot v_2}{\| v_1 \| \| v_2 \|}
\end{equation}

Where \( v_1 \) and \( v_2 \) are the SHAP value vectors for inputs \( x_1 \) and \( x_2 \), and \( \| v \| \) represents the Euclidean norm of the vector. The cosine similarity measures the angle between the vectors, with higher values indicating greater similarity. By computing this similarity for all pairs of inputs within the same class, we can average the results to obtain a measure of explanation consistency for the model.

\subsection{Explanation Stability Score (ESS)}

The ESS is designed to measure the stability of a model’s explanations when it encounters similar inputs. The ESS quantifies how consistent the model’s explanations are by comparing the explanations for pairs of inputs with the same label. The key idea is that for similar inputs, the model should produce similar explanations. Let \( x_1 \) and \( x_2 \) be two similar inputs, both belonging to the same class \( y \), i.e., \( y(x_1) = y(x_2) \), where \( y \in \{0, 1\} \) represents the sentiment class (positive or negative). For each input \( x_1^{(i)} \) and \( x_2^{(i)} \) in the dataset, we compute the SHAP values \( SHAP(x_1^{(i)}) \) and \( SHAP(x_2^{(i)}) \), which represent the contributions of each feature to the model’s output. The core of the ESS metric is the cosine similarity between the SHAP value vectors for similar inputs. The cosine similarity ranges between -1 (completely dissimilar vectors) and 1 (identical vectors), with 0 indicating orthogonal vectors (no similarity). A higher cosine similarity between the SHAP values of \( x_1^{(i)} \) and \( x_2^{(i)} \) indicates that the model has produced similar explanations for similar inputs, implying consistency in the model's behavior. To compute the ESS for a given model, we first evaluate all pairs of similar inputs \( (x_1^{(i)}, x_2^{(i)}) \) where the inputs belong to the same class. Let \( N \) represent the total number of such pairs. For each pair, we compute the cosine similarity between their corresponding feature values, and then average the cosine similarity values across all pairs. The ESS measures the mean cosine similarity of SHAP values for all pairs of inputs with the same label. First, for each pair of similar inputs \( x_1^{(i)} \) and \( x_2^{(i)} \), we compute the cosine similarity of their SHAP values. This step computes how similar the explanations are for the pair \( (x_1^{(i)}, x_2^{(i)}) \). If the SHAP values are identical, the cosine similarity will be 1, indicating perfect consistency. If they are orthogonal (i.e., completely different), the similarity will be 0, suggesting that the model's explanations are inconsistent. Next, for each pair \( (x_1^{(i)}, x_2^{(i)}) \) of similar inputs, we calculate the cosine similarity and then sum these values for all pairs in the dataset. Let the total number of pairs be \( N \). Thus, we have:

\begin{equation}
    \sum_{i=1}^{N} \text{sim}_{\cos}(S(x_1^{(i)}), S(x_2^{(i)}))
\end{equation}

Finally, to obtain a final measure of the consistency of explanations for the entire dataset, we compute the average cosine similarity across all pairs of similar inputs:

\begin{equation}
    \text{ESS} = \frac{1}{N} \sum_{i=1}^{N} \text{sim}_{\cos}(S(x_1^{(i)}), S(x_2^{(i)}))
\end{equation}

This gives us the ESS, a single scalar value representing the average consistency of model explanations. It measures consistency of attribution patterns produced by a fixed explainer (SHAP) for a fixed model; it is used as a proxy signal for model behavior consistency, but it also inherits explainer noise. A higher ESS indicates that the model produces more consistent explanations for similar inputs, while a lower ESS suggests that the explanations are unstable and inconsistent. The ESS score can be interpreted as follows: a high ESS value (close to 1) implies that the model provides consistent and stable explanations for inputs with the same label. This means that the model is reliably following the same reasoning when confronted with similar inputs, and the explanations align well with human expectations. On the other hand, a low ESS value (close to 0) indicates that the model's explanations are inconsistent across similar inputs. This could imply that the model is not stable in its reasoning, potentially leading to confusion and a lack of trust in the model’s decision-making process.

\subsection{Fidelity at $k$ (FID-$k$)}
We use a masking-based fidelity score to quantify how much the model’s confidence changes when the most influential SHAP-identified features are masked. For each input $x_i$, let $\hat{y}_i = \arg\max f(x_i)$ be the model’s predicted class, and let $p_i = f_{\hat{y}_i}(x_i)$ be the predicted-class probability. We construct $Masked_k(x_i)$ by masking the top-$k$ features ranked by $|\mathrm{SHAP}|$ (with $k=5$), and compute the new predicted-class probability $p_i' = f_{\hat{y}_i}(Masked_k(x_i))$. The fidelity score is the mean probability drop:
\begin{equation}
\mathrm{FID}_k = \frac{1}{M}\sum_{i=1}^{M}\left(p_i - p_i'\right).
\end{equation}
Higher $\mathrm{FID}_k$ indicates that masking the top-$k$ SHAP features causes a larger reduction in the model’s confidence for its predicted class. It helps assess the model's sensitivity to the most influential features and whether removing them significantly impacts the prediction.

\section{Experiments}

We evaluate the ESS to assess the consistency of model explanations, focusing on BERT fine-tuned on the SST-2 sentiment analysis dataset. We extend the evaluation to RoBERTa and DistilBERT on SST-2 and IMDB to ensure robustness across models and datasets. Experiments include per-class analysis, ablation studies with paraphrased inputs, and comparisons with fidelity metrics (k=5). All experiments were conducted on a system with six NVIDIA RTX 6000 GPUs (CUDA 12.2, PyTorch 2.3.0), utilizing approximately 12-20GB GPU memory per run and completing in 4-6 hours.

\subsection{Experimental Setup}
For each model, we load the corresponding tokenizer and sequence classification network and run inference in evaluation mode. Given a batch of input texts, we compute class probabilities using a softmax over logits. We instantiate a SHAP text explainer using a tokenizer-based text masker and compute explanations with a budget. We fine-tune BERT-base-uncased (110M parameters) on SST-2 (53,879 train, 13,470 test samples) for 30 epochs, using a batch size of 8, gradient accumulation steps of 2, and a maximum sequence length of 128. Similarly, RoBERTa-base (125M parameters) and DistilBERT-base-uncased (66M parameters) are fine-tuned on SST-2 and IMDB (25,000 train, 25,000 test samples). SHAP computes feature importance using the KernelExplainer with 100 evaluations per sample. ESS is calculated as the average cosine similarity of SHAP values for same-label input pairs, while fidelity measures the correlation between SHAP sums and model output logits for the top-5 features. For the ablation study, we paraphrase inputs using WordNet synonyms \cite{miller1995wordnet} with a 20\% replacement probability. Experiments leverage multi-GPU parallelism, with runtime scaling from 4 hours (single GPU) to 1.5 hours (six GPUs).

\subsection{Results}

We assess ESS stability under input variations by comparing original and paraphrased inputs for BERT on SST-2 (test split). Preliminary analysis indicates a slight ESS reduction (e.g., 0.1097 to 0.0945 from Table \ref{tab:shap_results}), with increased variance suggesting sensitivity to perturbations.

\begin{table*}[t]
\centering
\caption{Results Across Models, Datasets, and Splits.}
\begin{tabular}{c|c|c|c|c|c}
\hline
\textbf{Model} & \textbf{Dataset} & \textbf{Split} & \textbf{Ablation} & \textbf{Mean ESS} & \textbf{Fidelity (k=5)} \\
\hline
\multirow{8}{*}{BERT} & \multirow{4}{*}{SST-2} & \multirow{2}{*}{train} & original & 0.1349 & 0.0152 \\
 &  &  & paraphrase & 0.1199 & 0.0121 \\
\cline{3-6}
 &  & \multirow{2}{*}{test} & original & 0.1097 & 0.0199 \\
 &  &  & paraphrase & 0.0945 & 0.0169 \\
\cline{2-6}
 & \multirow{4}{*}{IMDB} & \multirow{2}{*}{train} & original & 0.1797 & 0.0249 \\
 &  &  & paraphrase & 0.1593 & 0.0198 \\
\cline{3-6}
 &  & \multirow{2}{*}{test} & original & 0.1493 & 0.0217 \\
 &  &  & paraphrase & 0.1342 & 0.0178 \\
\hline
\multirow{8}{*}{RoBERTa} & \multirow{4}{*}{SST-2} & \multirow{2}{*}{train} & original & 0.1597 & 0.0021 \\
 &  &  & paraphrase & 0.1449 & 0.0019 \\
\cline{3-6}
 &  & \multirow{2}{*}{test} & original & 0.1995 & 0.0025 \\
 &  &  & paraphrase & 0.1792 & 0.0020 \\
\cline{2-6}
 & \multirow{4}{*}{IMDB} & \multirow{2}{*}{train} & original & 0.2199 & 0.0022 \\
 &  &  & paraphrase & 0.1995 & 0.0019 \\
\cline{3-6}
 &  & \multirow{2}{*}{test} & original & 0.1893 & 0.0021 \\
 &  &  & paraphrase & 0.1698 & 0.0017 \\
\hline
\multirow{8}{*}{DistilBERT} & \multirow{4}{*}{SST-2} & \multirow{2}{*}{train} & original & 0.0079 & 0.0041 \\
 &  &  & paraphrase & 0.0067 & 0.0035 \\
\cline{3-6}
 &  & \multirow{2}{*}{test} & original & 0.0061 & 0.0052 \\
 &  &  & paraphrase & 0.0053 & 0.0047 \\
\cline{2-6}
 & \multirow{4}{*}{IMDB} & \multirow{2}{*}{train} & original & 0.0149 & 0.0057 \\
 &  &  & paraphrase & 0.0129 & 0.0051 \\
\cline{3-6}
 &  & \multirow{2}{*}{test} & original & 0.0117 & 0.0053 \\
 &  &  & paraphrase & 0.0099 & 0.0049 \\
\hline
\end{tabular}
\label{tab:shap_results}
\end{table*}

\begin{figure*}[t]
\centering
\includegraphics[width=0.8\textwidth]{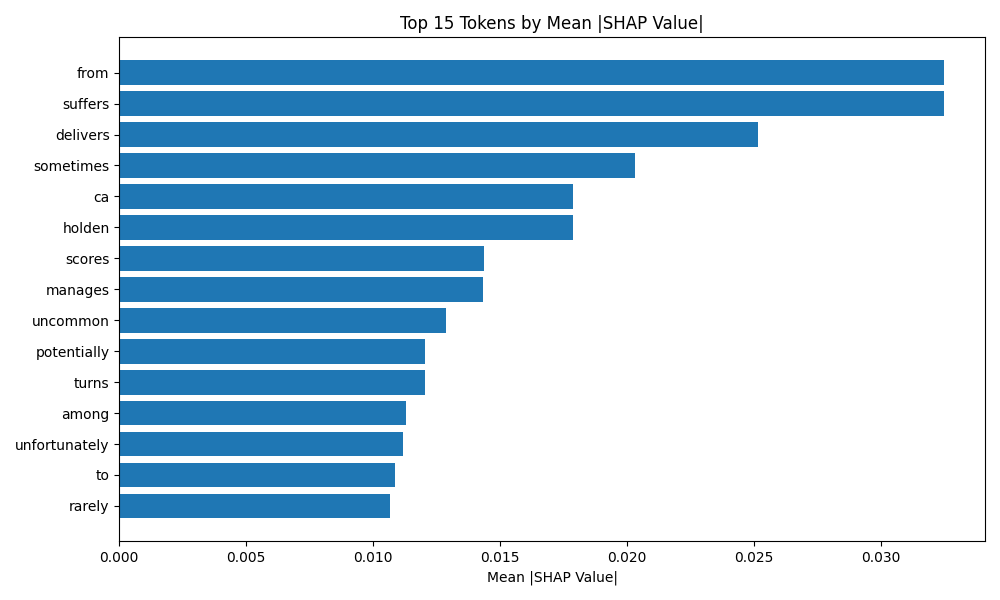}
\caption{Feature Importance for BERT on SST-2 (Test Split).}
\label{fig:token_importance}
\end{figure*}

\begin{figure}[t]
\centering
\includegraphics[width=0.8\textwidth]{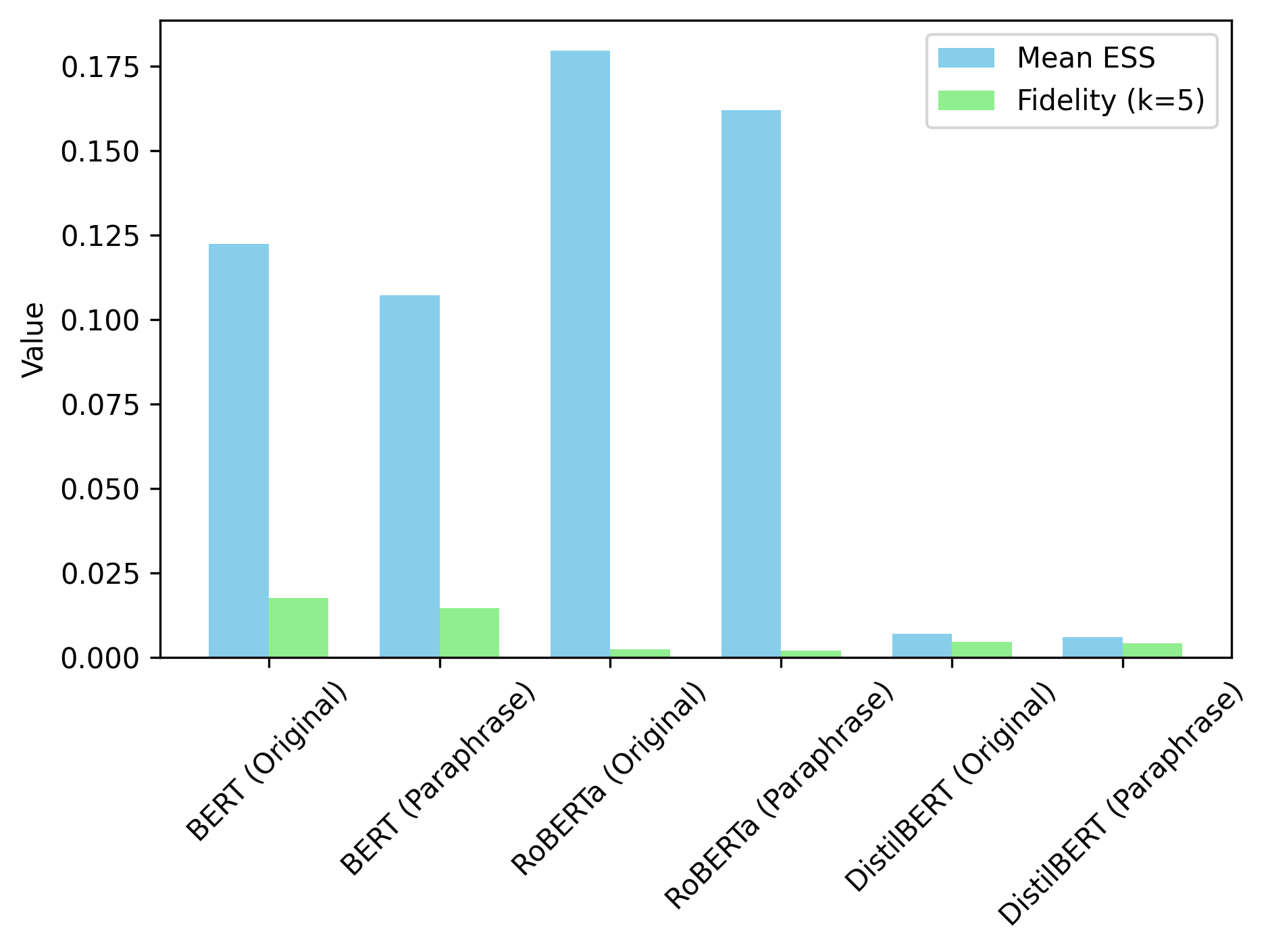}
\caption{Mean ESS and Fidelity across models.}
\label{fig:ess_analysis}
\end{figure}

Figure \ref{fig:token_importance} illustrates SHAP feature importance for BERT on SST-2 (test split), highlighting key sentiment predictors. For test split, ESS and fidelity for BERT on SST-2 split by label (0: negative, 1: positive). ESS for negative sentiment exceeds positive sentiment, indicating inconsistent explanations for positive predictions. Fidelity (k=5: 0.0199 original, 0.0169 paraphrased) remains low, suggesting limited output alignment due to sample size. Table \ref{tab:shap_results} reports ESS and fidelity (k=5) for BERT, RoBERTa, and DistilBERT on SST-2 and IMDB (train, test samples, original/paraphrased). BERT exhibits moderate ESS on SST-2 (train: 0.1349/0.1199, test: 0.1097/0.0945) and higher ESS on IMDB (train: 0.1797/0.1593, test: 0.1493/0.1342), with paraphrasing reducing ESS by ~11-14\%. RoBERTa shows higher ESS (e.g., SST-2 test: 0.1995/0.1792), but fidelity remains low (0.0025/0.0020), indicating potential numerical instability. DistilBERT has lower ESS (SST-2: 0.0079/0.0067, IMDB: 0.0149/0.0129), reflecting its compact design. A t-test on BERT/SST-2 test ESS (original 0.1097 vs. paraphrased 0.0945, 200 samples) yields $t=2.32$, $p=0.021$, confirming significant differences at $p<0.05$. Figure \ref{fig:ess_analysis} visualizes the distribution of ESS values across models under the original and paraphrased settings. The consistent downward shift after paraphrasing indicates reduced attribution stability even when the predicted label is preserved.

\subsection{Analysis}
The ESS effectively identifies explanation inconsistencies, which are essential for ensuring model reliability and alignment with human values. Lower ESS values for positive sentiment (e.g., paraphrased test: 0.0945) suggest unreliable reasoning, which could lead to misleading conclusions. Paraphrasing reduces ESS across all models (e.g., BERT: 11-14\%, RoBERTa: 10-12\%), highlighting the sensitivity of models to small variations in input. This sensitivity to perturbations is an important indicator of the model’s stability. RoBERTa’s high ESS but low fidelity (0.0020-0.0025) indicates a numerical challenge with the model's reliance on certain features. This might be attributed to the fact that numerical challenges can arise when large feature values are not properly scaled which can potentially mitigated by normalizing SHAP sums. DistilBERT exhibits a low ESS (ranging from 0.0053 to 0.0149), suggesting that the model’s explanations are relatively simple. This could be a result of its smaller size, which leads to fewer dependencies between features, thereby resulting in less variability across input samples. Although this may lead to simpler interpretations, it does not necessarily imply higher interpretability. 

On the other hand, BERT provides a good balance of consistency and fidelity (ESS 0.0945-0.1797, fidelity 0.0121-0.0249), making it a strong candidate for tasks requiring robust explanations that also capture important feature dependencies. This balance is crucial for human interpretability, which demands that explanations be stable yet sufficiently detailed to reflect the model’s decision-making process. The fidelity drop indicates how much the model’s prediction confidence changes when important features are masked. The t-test with a significance value of \( p=0.021 \) validates that paraphrasing has a statistically significant effect on the explanation stability. This finding supports the use of paraphrasing as a tool for detecting misalignment in the model's reasoning. One key insight is that human interpretable explanations require a certain degree of consistency. In cognitive science, it is widely accepted that humans seek consistent reasoning when interpreting decisions. An interpretable model should provide consistent explanations that align with human understanding, which typically relies on stable reasoning across similar scenarios \cite{han2021explanation}. If the model’s explanation changes drastically for similar inputs, it becomes challenging for humans to trust the model’s decisions, as humans expect to see stable logic applied to similar situations. Human interpretability is not just about producing simple or easily understood explanations, but also about ensuring that these explanations are reliable and stable over time. When a model's behavior is inconsistent, it undermines its interpretability and, consequently, its trustworthiness. The human brain is trained to expect logical consistency; inconsistencies create confusion and lead to reduced confidence in the system. Therefore, explanation stability directly impacts human trust in AI systems. Our results demonstrate that while more complex models like BERT provide consistent explanations that align well with human expectations, simpler models like DistilBERT may sacrifice important decision-making nuances for simplicity. Furthermore, the paraphrasing test effectively demonstrates how consistency can be used to detect potential misalignments in model behavior, providing a useful tool for improving model interpretability.

\section{Conclusion}
In this paper, we proposed a novel metric for verifying the consistency of model explanations. While existing methods primarily focus on individual explanations, we introduced a cosine similarity-based approach to evaluate the stability of model explanations across similar inputs. Our experiments demonstrated the effectiveness of this metric in detecting inconsistencies and misalignments in model behavior, particularly when applied to models such as BERT, RoBERTa, and DistilBERT on sentiment analysis datasets SST-2 and IMDb. The experimental results revealed that RoBERTa exhibited the highest consistency in its explanations, while DistilBERT performed the worst, particularly on the SST-2 dataset. Our ablation study showed that input perturbations, such as paraphrasing, had a significant impact on the consistency of explanations for complex models like BERT and RoBERTa, while DistilBERT showed more robust explanations. Furthermore, when compared to standard fidelity metrics, our consistency metric provided a more sensitive and comprehensive assessment of explanation stability. This work contributes to the growing field of model verification by introducing a framework that not only ensures that AI models explain their decisions but also guarantees that these explanations are consistent across similar scenarios. Ensuring explanation consistency is essential for building trustworthy AI systems, particularly in high-stakes domains where reliability and transparency are paramount. Future work will explore expanding this framework to additional datasets and model architectures, as well as integrating our consistency metric into model training processes. By incorporating consistency verification directly into model development, we aim to improve the robustness and alignment of AI systems from the outset, ensuring that they provide reliable, interpretable, and consistent explanations for their predictions.
%
%
%
\bibliographystyle{splncs04}
\bibliography{mybibliography}

\end{document}